\pdfoutput=1

\documentclass[11pt, final]{article}

\usepackage[final]{acl}
\usepackage{pdfpages}

\usepackage{times}
\usepackage{latexsym}
\usepackage{amsmath}

\usepackage[T1]{fontenc}
\usepackage[T5]{fontenc}

\usepackage[utf8]{inputenc}

\usepackage{microtype}

\usepackage{inconsolata}

\title{Coreference Resolution for Vietnamese Narrative Texts}

\author{Hieu-Dai Tran\textsuperscript{1, 2}, Duc-Vu Nguyen\textsuperscript{1, 2}, Ngan Luu-Thuy Nguyen\textsuperscript{1, 2} \\
\textsuperscript{1}University of Information Technology, Ho Chi Minh City, Vietnam\\
\textsuperscript{2}Vietnam National University, Ho Chi Minh City, Vietnam \\
\texttt{daith.15@grad.uit.edu.vn}\quad
\texttt{\{vund, ngannlt\}@uit.edu.vn}}

\begin{document}
\maketitle
\begin{abstract}
Coreference resolution is a vital task in natural language processing (NLP) that involves identifying and linking different expressions in a text that refer to the same entity. This task is particularly challenging for Vietnamese, a low-resource language with limited annotated datasets. To address these challenges, we developed a comprehensive annotated dataset using narrative texts from VnExpress, a widely-read Vietnamese online news platform. We established detailed guidelines for annotating entities, focusing on ensuring consistency and accuracy. Additionally, we evaluated the performance of large language models (LLMs), specifically GPT-3.5-Turbo and GPT-4, on this dataset. Our results demonstrate that GPT-4 significantly outperforms GPT-3.5-Turbo in terms of both accuracy and response consistency, making it a more reliable tool for coreference resolution in Vietnamese.
\end{abstract}

\section{Introduction}
Entity coreference resolution is a critical task in NLP that involves identifying and linking various expressions in a text that refer to the same entity \citep{jurafsky2014speech, ng2010machine, pradhan2011conll}. This task is essential for improving the coherence and understanding of texts in applications such as machine translation \citep{mitkov1998corpus}, information extraction \citep{grishman1997information}, and text summarization \citep{steinberger2007text}. Significant achievements have been made in coreference resolution, particularly for the English language, where numerous models and annotated datasets such as OntoNotes and CoNLL-2012 have been developed \citep{pradhan2012conll, hovy2006ontonotes}. Techniques range from early machine learning approaches to more recent neural network models \citep{lee2017end, clark2016deep}. However, coreference resolution for Vietnamese is still in its developmental stages, primarily due to the lack of comprehensive annotated datasets. As highlighted in \citet{hoang2023vihos}, the development of high-quality annotated datasets for Vietnamese NLP tasks is still an ongoing challenge. The ViHOS dataset, for instance, was created to address this gap in hate and offensive speech detection, indicating the broader need for such resources across various NLP tasks.

LLMs such as GPT-3.5-Turbo and GPT-4 have shown great promise across various domains in NLP, particularly in tasks involving zero-shot and few-shot learning \citep{brown2020language, radford2019language}. These models can leverage large amounts of data and transfer learning capabilities to perform well even with limited task-specific data. This makes LLMs excellent candidates for exploring tasks like coreference resolution in low-resource languages such as Vietnamese.

Evaluating the performance of LLMs in resolving coreference is an intriguing area of research. With the proliferation of various LLMs, there is substantial potential to explore and benchmark their capabilities in different contexts. Our study aims to address the gap in Vietnamese coreference resolution by leveraging the power of LLMs.

\begin{figure*}[htbp]
    \centering
    \includegraphics[width=\textwidth]{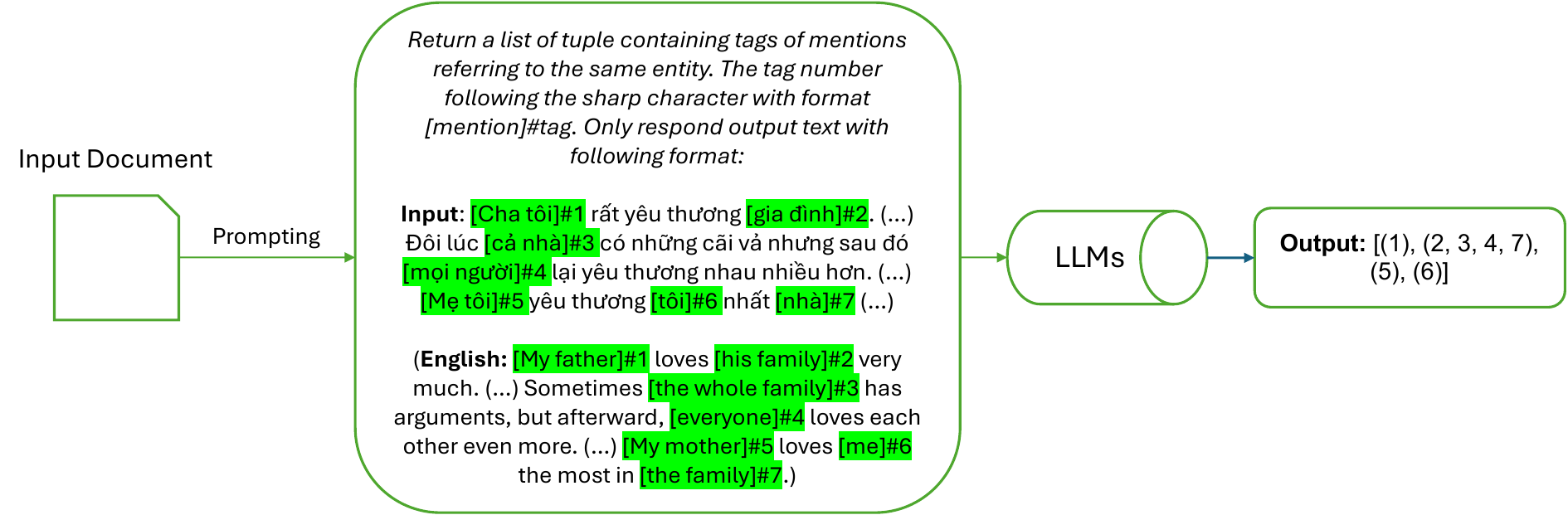}
    \caption{The process of generating mention clusters from raw text using LLMs. The input document is processed to return a list of tuples containing tags of mentions referring to the same entity. For example, in the input text, tags identify various entities and group them into clusters, as shown in the output.}
\end{figure*}

In this research, we collected a dataset from VnExpress, a popular Vietnamese online news platform, encompassing a wide range of narrative texts covering topics such as relationships, daily life, work, and social connections. We established detailed guidelines for annotating entities within these texts and carried out the annotation process manually. Furthermore, we used prompts to extract annotated entities from LLMs and evaluated their performance against our manually annotated dataset.

Our contributions are as follows: (1) we provide a comprehensive annotated dataset of Vietnamese narrative texts, (2) we develop detailed guidelines for entity annotation, (3) we use prompts to obtain annotated entities from LLMs, and (4) we evaluate the performance of LLMs against our annotated dataset to identify the most effective model for Vietnamese coreference resolution.

The rest of the paper is organized as follows: Section 2 reviews related work, Section 3 describes the dataset and annotation guidelines, Section 4 presents the evaluation of LLMs on the dataset, and Section 5 concludes the paper with future directions.

\section{Dataset Construction}
\subsection{Collecting Narrative Texts}
The data for our research was collected from VnExpress, a prominent Vietnamese online news platform, which has been a valuable source for various NLP tasks due to its rich narrative content \citep{nguyen2018vncore}. Similar efforts to use narrative texts for coreference tasks have been seen in other low-resource languages such as Hindi \citep{rahman2012low}.

\begin{table*}[h]
\centering
\begin{tabular}{|l|c|c|c|c|}
\hline
 & \textbf{Total} & \textbf{Average length} & \textbf{Average mention} & \textbf{Average entity} \\ \hline
\textbf{Few-shot} & 3 & 248.6 & 31 & 8.6 \\ \hline
\textbf{Evaluation} & 263 & 449.5 & 55.1 & 9.4 \\ \hline
\end{tabular}
\caption{The total dataset is divided into two subsets: an evaluation dataset consisting of 263 text files, and a few-shot dataset containing 3 text files.}
\label{tab:data_collection}
\end{table*}

Originally, the dataset consisted of 1,041 narrative texts, as described in the paper by \citet{nguyen2023abusive}. The texts are categorized into abusive and non-abusive to identify whether a text contains abusive content. For the purposes of our research, we randomly chose 266 texts from this original dataset. Table~\ref{tab:data_collection} shows the breakdown of the total dataset into evaluation and few-shot sets.

All selected texts were in their raw form, devoid of any prior annotations. In the following section, we outline the guidelines used to systematically annotate these texts.

\subsection{Annotation Guidelines}

In this section, we outline the guidelines and tools used for annotating entities within the selected narrative texts. The annotation process is crucial for ensuring consistency and accuracy in the data, which will be used for coreference resolution tasks. Annotation consistency is critical for the quality of coreference resolution datasets, as highlighted in previous studies \citep{poesio2018anaphora}. The annotation step was handled manually with the assistance of volunteers. Each volunteer was provided with the guidelines to ensure consistent annotation across the dataset.

\subsubsection{Tool for Annotation}

We utilized the open-source tool, Coreference Annotation Tool with SACR \citep{oberle2018sacr}. This tool facilitates faster and more intuitive annotation of entities. It features an intuitive user interface by coloring mentions referring to the same entity with a distinct color, making it easier to identify and annotate entities consistently. Each entity is assigned a unique color. Additionally, the tool supports annotating nested mentions, which is particularly helpful in cases of possessive mentions. After annotation, the tool exports the original text with entities wrapped in the format \{M\{tag\_number\} entity\_name\}. Entities referring to the same entity will share the same tag\_number.

\begin{figure*}[htbp]
    \centering
    \includegraphics[width=\textwidth]{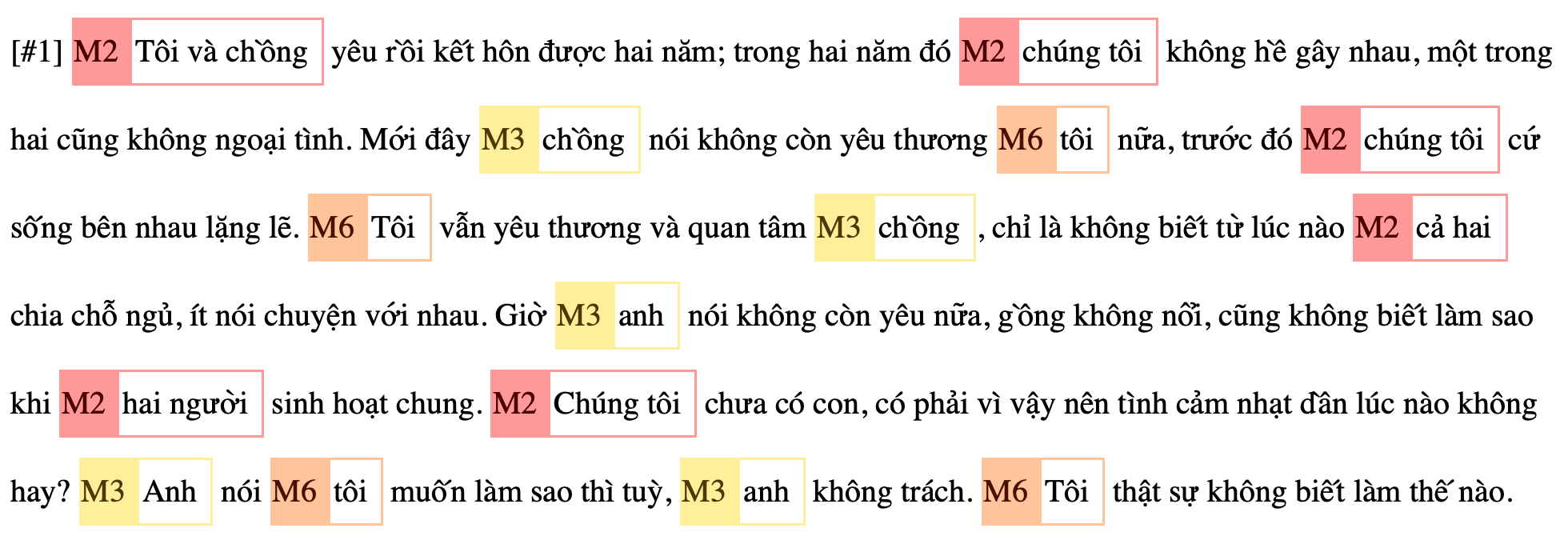}
    \caption{User interface of the Coreference Annotation Tool with SACR}
    \label{fig:annotation_tool}
\end{figure*}

\subsubsection{Definition of Entities}

In our research, we focused on annotating entities that refer to people, excluding non-human entities. This includes individuals, groups of people, organizations, or any reference to humans. We adhered to a set of simple rules for annotating these entities to maintain clarity and consistency:

\begin{enumerate}
    \item \textbf{People Mentions Only}: We annotate only those mentions that refer to entities which are human. We do not consider mentions that are nouns or names but are not human, such as geopolitical entities, objects, or places. For example:
    \begin{quote}
    Original: ``Anh ta chăn nuôi vịt ở Thái Lan.''\\
    (English: ``He raises ducks in Thailand'')\\
    Annotated: ``\textbf{\{M1 Anh ta\}} chăn nuôi vịt ở Thái Lan.''\\
    (English: ``\textbf{\{M1 He\}} raises ducks in Thailand'')
    \end{quote}

    \item \textbf{Groups of People}: Mentions that refer to organizations like WHO, or the Vietnam government, can also be considered groups of people, so we annotate those mentions as well. For example:
    \begin{quote}
    Original: WHO đang làm việc trên một sáng kiến y tế mới.\\
    (English: ``WHO is working on a new health initiative.'')\\
    Annotated: ``\textbf{\{M1 WHO\}} đang làm việc trên một sáng kiến y tế mới.''\\
    (English: ``\textbf{\{M1 WHO\}} is working on a new health initiative.'')
    \end{quote}
    
    \item \textbf{Excluding Adjectives}: When an entity includes adjectives, we annotate only the root noun without the adjective. For example:
    \begin{quote}
    Original: ``Chàng trai cao ráo khiến mọi người phải ngước nhìn.''\\
    (English: ``The tall guy makes everyone turn their heads.'')\\
    Annotated: ``\textbf{\{M1 Chàng trai\}} cao ráo luôn khiến \textbf{\{M2 mọi người\}} phải ngước nhìn.''\\
    (English: ``The tall \textbf{\{M1 guy\}} makes \textbf{\{M2 everyone\}} turn their heads.'')\\
    \end{quote}

    \item \textbf{Nested Mentions}: Nested mentions are those that exist within another mention. We only annotate nested mentions when they are in an explicit possessive form, which includes the word ``của'' (meaning ``of'' in English) in the mention. For example:
    \begin{quote}
    Original: ``Mẹ của tôi thường thức dậy vào lúc 5 giờ sáng.''\\
    (English: ``My mother usually wakes up at 5 a.m.'')\\
    Annotated: ``\textbf{\{M1 Mẹ của \textbf{\{M2 tôi\}}\}} thường thức dậy vào lúc 5 giờ sáng.''
    (English: ``\textbf{\{M1 My mother}\} usually wakes up at 5 a.m.'')
    \end{quote}
    In this case, \textbf{\{M2 tôi\}} is the nested mention within \textbf{\{M1 Mẹ\}}, and they are separated by the word ``của.''

    \textbf{Note:} The nested mention \textbf{\{M2 tôi\}} represents ``I'' in the English translation and is explicitly shown as a nested entity in Vietnamese. However, this nested structure cannot be represented as explicitly in English, which leads to discrepancies in how nested entities are handled between the two languages. This difference poses challenges for cross-linguistic coreference resolution tasks.

\end{enumerate}

These guidelines help ensure that our annotations are focused on the relevant entities for coreference resolution, making the data more useful for subsequent analysis and model training.

\section{Prompting}
Prompting is a process where we design a prompt to request a response from the LLMs and receive the response back. This technique, especially in the context of few-shot learning, has been shown to significantly improve performance in various NLP tasks \citep{brown2020language, gao2020making}. The effectiveness of prompts in coreference resolution tasks has been discussed in \citep{liu2021pre}, demonstrating their utility in scenarios with limited training data. This step is time-consuming as it often requires trying many different prompts to achieve the desired response. The role of prompting, especially in the context of few-shot learning, has been extensively discussed in recent research \citep{wei2021prompt}.

In this research, we employed few-shot learning by providing the LLMs with some examples of input and output, then asking them to respond to new inputs. This method helps the LLMs understand not only the context of the query but also the format in which the response should be provided. Few-shot learning has shown significant promise in enhancing the performance of language models across various tasks \citep{brown2020language, lin2021fewshot}.

Specifically, we took 3 texts annotated from the raw dataset and used them to build the prompt as examples for the LLMs. These 3 texts covered the rules discussed in the ``Definition of Entities'' section as much as possible with a reasonable amount. This left us with 263 texts to verify, comparing the output we annotated manually against the output generated by the LLMs. The reason why we only took 3 texts is that some LLMs, like GPT-4, accept a limited number of tokens per request. This limitation is a constraint when using the API, and we must ensure that the prompt and expected output stay within this limit.

Before building the prompt, we also had to take a few steps to refine the annotated dataset to build the \textit{gold\_clusters}, which is the expected result that we want the LLMs to return. First, we'll format the text annotated with the format \textbf{\{M\{\#tag\_number\} mention\}} to \textbf{[mention]\{\#tag\_index\}}. Here is an example:

\begin{quote}
Annotated: \textbf{\{M1 Em\}} trân trọng \textbf{\{M2 hai người bạn\}} rất thân; \textbf{\{M2 các bạn\}} bị \textbf{\{M3 một nhóm bạn khác\}} nói xấu rất nhiều, từ tính cách, lời nói, dáng đi. \textbf{\{M1 Em\}} chắc chắn trước đó \textbf{\{M2 hai bạn\}} không đả động gì tới \textbf{\{M3 nhóm bạn đó\}}...

(English: \textbf{\{M1 I\}} deeply value \textbf{\{M2 my two close friends\}}; \textbf{\{M2 they\}} are being talked about negatively by \textbf{\{M3 another group of friends\}}, criticizing everything from their personality, speech, to their posture. \textbf{\{M1 I\}}'m certain that before this, \textbf{\{M2 my two friends\}} hadn't done or said anything to \textbf{\{M3 that group\}}.)

Indexed: \textbf{[Em]\#1} trân trọng \textbf{[hai người bạn]\#2} rất thân; \textbf{[các bạn]\#3} bị \textbf{[một nhóm bạn khác]\#4} nói xấu rất nhiều, từ tính cách, lời nói, dáng đi. \textbf{[Em]\#5} chắc chắn trước đó \textbf{[hai bạn]\#6} không đả động gì tới \textbf{[nhóm bạn đó]\#7}...

(English: \textbf{[I]\#1} deeply value \textbf{[my two close friends]\#2}; \textbf{[they]\#3} are being talked about negatively by \textbf{[another group of friends]\#4}, criticizing everything from their personality, speech, to their posture. \textbf{[I]\#5}'m certain that before this, \textbf{[my two friends]\#6} hadn't done or said anything to \textbf{[that group]\#7}.)

Then the \textit{gold\_clusters} should be: [(1, 5), (2, 3, 6), (4, 7)]
\end{quote}

The \textit{gold\_clusters} is formatted as an array of tuples, such as [(1, 5), (2, 3, 6), (4, 7)], where 1 and 5 are \textit{tag\_indices} of mentions that refer to a single entity as illustrated in the \textit{indexed} and \textit{annotated} above. The indices 2, 3, and 6 refer to another distinct entity, while 4 and 7 correspond to yet another entity.

There are two key reasons for using this specific format for \textit{gold\_clusters}:

\begin{itemize}
    \item \textbf{Efficiency in LLM Processing}: By formatting the output as tuples representing \textit{tag\_indices}, we minimize the number of tokens the LLM needs to generate. This results in a faster response time, which is crucial when processing large datasets or when multiple iterations of prompting are necessary.

    \item \textbf{Seamless Integration into Evaluation}: The tuple-based format is directly aligned with the requirements of our evaluation step. By receiving the output in this format, we can bypass additional processing steps that would otherwise be needed if the full text were returned. This streamlined approach allows us to directly compute the differences between the LLMs' output and the \textit{gold\_clusters}, enhancing the efficiency and accuracy of our evaluation process.
\end{itemize}
These considerations make the tuple-based format an optimal choice for both the performance of the LLMs and the subsequent evaluation of their outputs, ultimately contributing to a more efficient and effective coreference resolution process.

After building the \textit{gold\_clusters} for the 3 documents we selected in the previous step, we were ready to construct the full prompt. To do this, we combined the \textit{indexed text} with the corresponding \textit{gold\_clusters}. The \textit{indexed text} served as the input, and the \textit{gold\_clusters} served as the output, formatted as follows: \textbf{Input: \{indexed text\} Output: \{gold\_clusters\}}. This constituted the few-shot learning part, which we presented to the LLMs for each document we wanted them to process. We refer to this part as the ``few-shot prompt.''

The next step involved processing each remaining document in the dataset (excluding the 3 documents used to build the few-shot prompt). We followed these steps:

\begin{enumerate}
    \item \textbf{Format the Document to Indexed Text}: Convert the document into the \textit{indexed text} format, as described previously.
    
    \item \textbf{Build the Final Prompt}: The final prompt began with the few-shot prompt, followed by a request for the LLMs to return the full output, with the input being the \textit{indexed text} created in step 1.
    
    \item \textbf{Send the Final Prompt to LLMs}: Submit the final prompt to the LLMs and save the resulting output for later evaluation against the corresponding \textit{gold\_clusters}.
\end{enumerate}

By following this systematic approach, we ensured that the LLMs were provided with consistent and well-structured prompts, which should enhance the accuracy and efficiency of the coreference resolution process. The outputs generated by the LLMs were stored and later compared with the manually annotated \textit{gold\_clusters} to evaluate their performance.

\section{Evaluation}

To evaluate the performance of the LLMs in coreference resolution for Vietnamese narrative texts, we conducted an experiment using two versions of OpenAI's GPT models. Similar evaluation methods using large-scale models for coreference have been employed in other studies, which leverage the CoNLL F1 score and its associated metrics like MUC, B-Cubed, and CEAF$_\phi$ \citep{pradhan2011conll, luo2005coreference}. The evaluation was carried out by comparing the outputs generated by these models against the manually annotated \textit{gold\_clusters}. The primary metrics used for this evaluation include the CoNLL F1 score, which is an aggregate of three metrics: MUC, B-Cubed, and CEAF$_\phi$.

\subsection{Evaluation Metrics}

The evaluation metrics used in this study are as follows:

\begin{itemize}
    \item \textbf{MUC (Mention-Pair)}: The MUC metric, introduced by \citet{vilain1995model}, evaluates the overlap between predicted and actual coreference clusters by considering links between mentions. The precision and recall for MUC are calculated as follows:

    \[
    \text{Precision} = \frac{L_{\text{correct}}}{L_{\text{predicted}}}
    \]
    \[
    \text{Recall} = \frac{L_{\text{correct}}}{L_{\text{gold}}}
    \]

    where \( L_{\text{correct}} \) is the number of correctly predicted links, \( L_{\text{predicted}} \) is the total number of predicted links, and \( L_{\text{gold}} \) is the total number of links in the gold standard. The MUC F1 score is the harmonic mean of precision and recall.

    \item \textbf{B-Cubed}: The B-Cubed metric, proposed by \citet{bagga1998algorithms}, evaluates coreference resolution at the mention level. For each mention, precision and recall are calculated as:

    \[
    \text{Precision} = \frac{|C_i \cap G_i|}{|C_i|}
    \]
    \[
    \text{Recall} = \frac{|C_i \cap G_i|}{|G_i|}
    \]

    where \( C_i \) is the set of mentions in the predicted cluster for mention \( i \), and \( G_i \) is the set of mentions in the gold cluster for mention \( i \). The overall B-Cubed precision and recall are averaged over all mentions, and the B-Cubed F1 score is the harmonic mean of these averaged values.

    \item \textbf{CEAF$_\phi$ (Constrained Entity Alignment F-Score)}: The CEAF$_\phi$ metric, discussed by \citet{luo2005coreference}, measures the similarity between predicted and actual entity clusters by finding an optimal one-to-one alignment between them. Precision and recall for CEAF$_\phi$ are calculated as:

    \[
    \text{Precision} = \frac{\sum_{i=1}^{n} \phi(C_i, G_i)}{\sum_{i=1}^{n} \phi(C_i, C_i)}
    \]
    \[
    \text{Recall} = \frac{\sum_{i=1}^{n} \phi(G_i, G_i)}{\sum_{i=1}^{n} \phi(G_i, G_i)}
    \]

    where \( \phi \) is a similarity function, typically the size of the intersection between clusters, and the sums are over the aligned cluster pairs. The CEAF$_\phi$ F1 score is computed as the harmonic mean of precision and recall.

    \item \textbf{CoNLL F1}: The CoNLL F1 score is the average of the F1 scores from the MUC, B-Cubed, and CEAF$_\phi$ metrics, providing an overall evaluation of the coreference resolution performance \citep{pradhan2011conll}.

    \item \textbf{Response Consistency}: During the response collection by calling OpenAI's API, we observed that the responses from GPT-4 were more consistent than those from GPT-3.5-Turbo. Specifically, the responses from GPT-3.5-Turbo often included unrelated parts along with the response, requiring additional refinement to extract the final result. Additionally, GPT-3.5-Turbo sometimes returned an unnecessarily annotated full text. In contrast, this problem occurred much less frequently with GPT-4, making it more reliable and reducing the need for post-processing.
\end{itemize}

\subsection{Results}

\begin{table}[h]
\centering
\begin{tabular}{|l|c|c|}
\hline
\textbf{Metric} & \textbf{GPT-3.5-Turbo} & \textbf{GPT-4} \\ \hline
\textbf{CoNLL F1} & 0.478 & 0.735 \\ \hline
\textbf{MUC F1} & 0.640 & 0.858 \\ \hline
\textbf{B-Cubed F1} & 0.474 & 0.723 \\ \hline
\textbf{CEAF$_\phi$ F1} & 0.321 & 0.625 \\ \hline
\end{tabular}
\caption{Comparison of GPT-3.5-Turbo and GPT-4 performance on coreference resolution.}
\label{tab:results}
\end{table}

The evaluation results demonstrate the effectiveness of GPT-4 over GPT-3.5-Turbo in Vietnamese coreference resolution across all metrics. The models were assessed using the CoNLL F1 score, which aggregates MUC, B-Cubed, and CEAF$_\phi$ metrics. Table~\ref{tab:results} summarizes the performance differences between the two models.

GPT-4 achieved a CoNLL F1 score of 0.735, showing a significant improvement over GPT-3.5-Turbo, which scored 0.478. This indicates that GPT-4 is considerably more effective in accurately linking mentions to the correct entities throughout the dataset. The MUC metric, which evaluates the overlap of predicted and actual coreference clusters, showed that GPT-4 performed exceptionally well with an F1 score of 0.858, compared to 0.640 for GPT-3.5-Turbo. These results suggest that GPT-4 is better at identifying and linking mentions that refer to the same entity, resulting in fewer errors related to missed or incorrect links.

For the B-Cubed metric, which is sensitive to mention-level errors, GPT-4 achieved a score of 0.723, significantly outperforming GPT-3.5-Turbo’s score of 0.474. This indicates that GPT-4 assigns individual mentions to the correct entity clusters more accurately. The CEAF$_\phi$ metric, which measures the alignment between predicted and actual entity clusters, further validated GPT-4’s capabilities with a score of 0.625, while GPT-3.5-Turbo scored much lower at 0.321. This result highlights GPT-4’s consistency and accuracy in entity clustering, aligning more closely with human-annotated gold standards.

Additionally, response consistency during the evaluation process favored GPT-4. GPT-3.5-Turbo responses often included irrelevant content or returned annotated full texts, requiring additional refinement. In contrast, GPT-4 demonstrated greater consistency, with fewer errors, making it a more reliable tool for coreference resolution tasks with minimal post-processing needed.

\subsection{Case Study}

In this case study, we identify specific instances where GPT-4 demonstrated superior coreference resolution capabilities compared to GPT-3.5-Turbo, based on the provided narrative text. These instances highlight the differences in handling entity references, contributing to GPT-4’s better performance across evaluation metrics.

\textbf{Case 1: Accurate Clustering of References to the Speaker}

\begin{itemize}
    \item \textbf{Example Text:} Mentions of the speaker [Tôi] (I) throughout the text, such as “Tôi 32 tuổi, lấy chồng được chín năm, có hai con gái, đang suy nghĩ việc bỏ chồng” (I am 32 years old, have been married for nine years, have two daughters, and am considering leaving my husband).
    \item \textbf{GPT-4:} Correctly grouped all references to the speaker into a single cluster, maintaining consistency. For example, it included mentions like [Tôi], [tôi], and other references to the speaker across the text into one coherent cluster.
    \item \textbf{GPT-3.5-Turbo:} Merged references to the speaker with unrelated entities such as the husband, resulting in a single, overly broad cluster. This mistake blurred the distinction between different characters, leading to lower precision and recall scores in MUC and B-Cubed metrics.
\end{itemize}

\textbf{Case 2: Differentiation Between the Speaker and the Husband}

\begin{itemize}
    \item \textbf{Example:} Mentions of the husband [chồng] (husband) and [anh] (he) as distinct from the speaker [Tôi] (I). In the sentence “Tôi cũng vay riêng 290 triệu đồng để trả nợ cho anh” (I also borrowed 290 million VND to pay off his debt), the speaker and her husband are clearly distinct entities.
    \item \textbf{GPT-4:} Successfully differentiated between the speaker and the husband, creating separate clusters for each. This accuracy ensured that references to [chồng] and [anh] were not confused with those referring to [Tôi].
    \item \textbf{GPT-3.5-Turbo:} Often failed to differentiate between these entities, merging them into a single cluster. This error indicates a lack of precision in entity resolution, which can affect the overall understanding of the text.
\end{itemize}

\textbf{Case 3: Handling of Family References and Relationships}

\begin{itemize}
    \item \textbf{Example:} Mentions involving family relationships, such as “[bố tôi] thấy hai vợ chồng không ổn định công việc” (my father saw that the couple was not stable in their work), where [bố tôi] (my father) and [vợ chồng] (the couple) refer to different entities.
    \item \textbf{GPT-4:} Accurately handled these family-related references, correctly clustering mentions of [bố tôi] separately from [vợ chồng], which denotes both the speaker and her husband.
    \item \textbf{GPT-3.5-Turbo:} Struggled to keep these distinctions clear, sometimes merging family-related terms incorrectly into broader clusters, reducing the specificity needed for accurate coreference resolution.
\end{itemize}

\textbf{Case 4: Treatment of Noun Phrases and Generic References}

\begin{itemize}
    \item \textbf{Example:} Generic references and noun phrases like [hai con gái] (two daughters) and [con cái] (children), which need to be associated accurately. In the sentence “bỏ chồng lại nghĩ đến con cái” (leaving my husband, I think of the children), references to the children need to be linked correctly.
    \item \textbf{GPT-4:} Effectively grouped these mentions, maintaining a clear cluster that includes all references to the speaker's children, such as [hai con gái] and [con cái].
    \item \textbf{GPT-3.5-Turbo:} Failed to consistently group these mentions, sometimes treating them as unrelated or merging them with other unrelated clusters. This led to inaccuracies in capturing the relationship dynamics within the narrative.
\end{itemize}

\subsection{Discussion}

The results clearly demonstrate that GPT-4 is superior to GPT-3.5-Turbo in performing coreference resolution on Vietnamese narrative texts, a finding that aligns with similar studies where advanced transformer-based models outperform earlier architectures \citep{devlin2018bert, lewis2020bart}. These findings reinforce the trend that larger, more sophisticated models offer improved capabilities in capturing the nuances of low-resource languages \citep{conneau2020unsupervised}. The improvement across all metrics can be attributed to the more advanced architecture and training data of GPT-4, which aligns with findings from earlier work on few-shot learning with large language models \citep{brown2020language}. This allows GPT-4 to better understand the complexities of coreference in a low-resource language like Vietnamese.

While both models showed some level of proficiency, the substantial gap in performance underscores the importance of using more advanced LLMs like GPT-4 for tasks that require a nuanced understanding of language. The evaluation also highlights the areas where further improvements are needed, such as better handling of difficult cases like extracting the exact noun from a complicated noun phrase or understanding the semantics to link the correct entity.

Overall, the use of LLMs in Vietnamese coreference resolution appears promising, with GPT-4 paving the way for more accurate and reliable models that can handle the intricacies of the Vietnamese language.

\section{Conclusion}

In this study, we explored the application of LLMs, specifically GPT-3.5-Turbo and GPT-4, for the task of coreference resolution in Vietnamese narrative texts. Coreference resolution, a critical component of NLP, involves identifying and linking various expressions in a text that refer to the same entity. This task is particularly challenging for low-resource languages like Vietnamese, where annotated datasets are scarce.

We utilized a dataset originally created by \citet{nguyen2023abusive}, which was collected from VnExpress and covers a diverse range of narrative topics. We developed detailed guidelines for annotating entities within this dataset and leveraged the few-shot learning capabilities of LLMs to design prompts that allowed these models to perform coreference resolution on the dataset. The evaluation of the models' outputs against the manually annotated \textit{gold\_clusters} provided insights into their effectiveness.

The results of our evaluation clearly demonstrate the superiority of GPT-4 over GPT-3.5-Turbo in resolving coreferences in Vietnamese texts. GPT-4 achieved a CoNLL F1 score of 0.735, significantly outperforming GPT-3.5-Turbo, which scored 0.478. This improvement was consistent across all metrics, including MUC, B-Cubed, and CEAF$_\phi$, indicating that GPT-4 is more adept at accurately identifying and linking mentions to the correct entities.

\subsection{Future Work}

While this research has made significant strides in improving coreference resolution for Vietnamese, several areas remain open for further exploration. One promising direction is the expansion of the annotated dataset. Increasing its size and diversity by incorporating more narrative genres, regional dialects, and contemporary language use could significantly enhance the robustness and generalizability of the models. Another important avenue is the fine-tuning of models on domain-specific texts, such as legal documents, medical records, or historical texts. This would require the development of specialized annotated datasets and evaluation metrics tailored to specific domains.

Future work could also focus on integrating coreference resolution with other NLP tasks, such as sentiment analysis, machine translation, and information extraction. This integration has the potential to create more holistic language understanding systems capable of handling complex, multi-faceted text analysis tasks. At the same time, the development of more efficient models is critical, particularly for reducing the significant computational costs associated with large-scale models like GPT-4. Techniques such as model distillation or pruning could be explored to achieve a balance between high accuracy and resource efficiency.

Additionally, exploring multilingual and cross-lingual models could leverage the linguistic similarities between Vietnamese and other Southeast Asian languages, potentially enhancing coreference resolution across multiple languages. Cross-lingual transfer learning techniques may prove especially valuable for improving performance in languages with even fewer resources than Vietnamese. The incorporation of external knowledge sources, such as structured databases or knowledge graphs, could also bolster model performance, particularly in handling entities underrepresented in training data.

Efforts to improve how models handle ambiguities in coreference resolution are equally critical. Challenges such as pronoun resolution or implied entity references require more sophisticated context-awareness mechanisms within the models. Lastly, developing user-interactive coreference resolution tools could add significant value in applications such as content creation, editing, and data analysis. These tools could allow users to guide or correct the resolution process in real-time while leveraging user feedback to continually refine model performance.

The success of LLMs like GPT-4 represents a significant step forward in coreference resolution for Vietnamese. This aligns with findings from other studies that demonstrate the versatility of LLMs across languages and tasks, even those with limited training data \citep{radford2019language, raffel2020exploring}. However, there remains substantial potential for further innovation, particularly in areas such as dataset expansion, domain adaptation, model efficiency, and cross-lingual applications. These future directions hold great promise for developing more accurate and reliable NLP systems that can better address the linguistic diversity and complexity of Vietnamese and other low-resource languages.

\section*{Acknowledgement}
This research was supported by The VNUHCM-University of Information Technology’s Scientific Research Support Fund.

\bibliography{custom}

\begin{thebibliography}{29}
\providecommand{\natexlab}[1]{#1}

\bibitem[{Bagga and Baldwin(1998)}]{bagga1998algorithms}
Amit Bagga and Breck Baldwin. 1998.
\newblock Algorithms for scoring coreference chains.
\newblock In \emph{The first international conference on language resources and evaluation workshop on linguistic coreference}, pages 563--566.

\bibitem[{Brown et~al.(2020)Brown, Mann, Ryder et~al.}]{brown2020language}
Tom~B. Brown, Benjamin Mann, Nick Ryder, et~al. 2020.
\newblock Language models are few-shot learners.
\newblock \emph{arXiv preprint arXiv:2005.14165}.

\bibitem[{Clark and Manning(2016)}]{clark2016deep}
Kevin Clark and Christopher~D. Manning. 2016.
\newblock Deep reinforcement learning for mention-ranking coreference models.
\newblock \emph{arXiv preprint arXiv:1609.08667}.

\bibitem[{Conneau et~al.(2020)Conneau, Khandelwal, Goyal, Chaudhary, Wenzek, Guzm{\'a}n, Grave, Ott, Zettlemoyer, and Stoyanov}]{conneau2020unsupervised}
Alexis Conneau, Kartikay Khandelwal, Naman Goyal, Vishrav Chaudhary, Guillaume Wenzek, Francisco Guzm{\'a}n, Edouard Grave, Myle Ott, Luke Zettlemoyer, and Veselin Stoyanov. 2020.
\newblock Unsupervised cross-lingual representation learning at scale.
\newblock In \emph{Proceedings of the 58th Annual Meeting of the Association for Computational Linguistics}, pages 8440--8451.

\bibitem[{Devlin et~al.(2018)Devlin, Chang, Lee, and Toutanova}]{devlin2018bert}
Jacob Devlin, Ming-Wei Chang, Kenton Lee, and Kristina Toutanova. 2018.
\newblock Bert: Pre-training of deep bidirectional transformers for language understanding.
\newblock \emph{arXiv preprint arXiv:1810.04805}.

\bibitem[{Gao et~al.(2020)Gao, Fisch, and Chen}]{gao2020making}
Tianyu Gao, Adam Fisch, and Danqi Chen. 2020.
\newblock Making pre-trained language models better few-shot learners.
\newblock \emph{arXiv preprint arXiv:2012.15723}.

\bibitem[{Grishman(1997)}]{grishman1997information}
Ralph Grishman. 1997.
\newblock Information extraction: Techniques and challenges.
\newblock \emph{International Summer School on Information Extraction}.

\bibitem[{Hoang et~al.(2023)Hoang, Luu, Tran, Van~Nguyen, and Nguyen}]{hoang2023vihos}
Phu~Gia Hoang, Canh~Duc Luu, Khanh~Quoc Tran, Kiet Van~Nguyen, and Ngan Luu-Thuy Nguyen. 2023.
\newblock Vihos: Hate speech spans detection for vietnamese.
\newblock \emph{arXiv preprint arXiv:2301.10186}.

\bibitem[{Hovy et~al.(2006)Hovy, Marcus, Palmer, Ramshaw, and Weischedel}]{hovy2006ontonotes}
Eduard Hovy, Mitchell Marcus, Martha Palmer, Lance Ramshaw, and Ralph Weischedel. 2006.
\newblock Ontonotes: The 90\% solution.
\newblock In \emph{Proceedings of the Human Language Technology Conference of the NAACL, Companion Volume: Short Papers}, pages 57--60.

\bibitem[{Jurafsky and Martin(2014)}]{jurafsky2014speech}
Daniel Jurafsky and James~H. Martin. 2014.
\newblock \emph{Speech and Language Processing}.
\newblock Prentice Hall.

\bibitem[{Lee et~al.(2017)Lee, He, Lewis, and Zettlemoyer}]{lee2017end}
Kenton Lee, Luheng He, Mike Lewis, and Luke Zettlemoyer. 2017.
\newblock End-to-end neural coreference resolution.
\newblock \emph{arXiv preprint arXiv:1707.07045}.

\bibitem[{Lewis et~al.(2020)Lewis, Liu, Goyal, Ghazvininejad, Mohamed, Levy, Stoyanov, and Zettlemoyer}]{lewis2020bart}
Mike Lewis, Yinhan Liu, Naman Goyal, Marjan Ghazvininejad, Abdelrahman Mohamed, Omer Levy, Ves Stoyanov, and Luke Zettlemoyer. 2020.
\newblock Bart: Denoising sequence-to-sequence pre-training for natural language generation, translation, and comprehension.
\newblock In \emph{Proceedings of the 58th Annual Meeting of the Association for Computational Linguistics}, pages 7871--7880.

\bibitem[{Lin et~al.(2021)Lin, Wu, Lee, Wang, and Ren}]{lin2021fewshot}
Bill~Yuchen Lin, Ziyi Wu, Sue Lee, Yichi Wang, and Xiang Ren. 2021.
\newblock Few-shot learning with multilingual generative language models.
\newblock \emph{arXiv preprint arXiv:2112.10668}.

\bibitem[{Liu et~al.(2021)}]{liu2021pre}
Pengfei Liu et~al. 2021.
\newblock Pre-train, prompt, and predict: A systematic survey of prompting methods in natural language processing.
\newblock \emph{arXiv preprint arXiv:2107.13586}.

\bibitem[{Luo(2005)}]{luo2005coreference}
Xiaoqiang Luo. 2005.
\newblock On coreference resolution performance metrics.
\newblock In \emph{Proceedings of the 2005 conference on empirical methods in natural language processing}, pages 25--32.

\bibitem[{Mitkov(1998)}]{mitkov1998corpus}
Ruslan Mitkov. 1998.
\newblock A corpus-based approach to pronoun resolution.
\newblock In \emph{Proceedings of the 17th international conference on Computational linguistics}.

\bibitem[{Ng(2010)}]{ng2010machine}
Vincent Ng. 2010.
\newblock Machine learning for coreference resolution: From local classification to global ranking.
\newblock \emph{Proceedings of the ACL}.

\bibitem[{Nguyen et~al.(2018)Nguyen, Nguyen, Nguyen, Nguyen, and Pham}]{nguyen2018vncore}
Dat~Quoc Nguyen, Dai~Quoc Nguyen, Dang-Khoa Le-Tuan Nguyen, Son~Bao Nguyen, and Son~T. Pham. 2018.
\newblock Vncorenlp: A vietnamese natural language processing toolkit.
\newblock In \emph{Proceedings of the 2018 Conference of the North American Chapter of the Association for Computational Linguistics: Demonstrations}, pages 56--60.

\bibitem[{Nguyen et~al.(2023)Nguyen, Phan, Nguyen, and Nguyen}]{nguyen2023abusive}
Nhu-Thanh Nguyen, Khoa Thi-Kim Phan, Duc-Vu Nguyen, and Ngan Luu-Thuy Nguyen. 2023.
\newblock \href {https://arxiv.org/abs/2312.07831} {Abusive span detection for vietnamese narrative texts}.
\newblock \emph{arXiv preprint arXiv:2312.07831}.

\bibitem[{Oberle(2018)}]{oberle2018sacr}
Benedikt Oberle. 2018.
\newblock \href {https://aclanthology.org/L18-1059} {Sacr: A drag-and-drop based tool for coreference annotation}.
\newblock In \emph{Proceedings of the Eleventh International Conference on Language Resources and Evaluation (LREC 2018)}. European Language Resources Association (ELRA).

\bibitem[{Poesio and Artstein(2008)}]{poesio2018anaphora}
Massimo Poesio and Ron Artstein. 2008.
\newblock Anaphora resolution: State of the art.
\newblock In \emph{Proceedings of the ACL}.

\bibitem[{Pradhan et~al.(2011)Pradhan, Ramshaw, Marcus, Palmer, Weischedel, and Xue}]{pradhan2011conll}
Sameer Pradhan, Lance Ramshaw, Mitch Marcus, Martha Palmer, Ralph Weischedel, and Nianwen Xue. 2011.
\newblock Conll-2011 shared task: Modeling unrestricted coreference in ontonotes.
\newblock In \emph{Proceedings of the Fifteenth Conference on Computational Natural Language Learning: Shared Task}, pages 1--27.

\bibitem[{Pradhan et~al.(2012)}]{pradhan2012conll}
Sameer Pradhan et~al. 2012.
\newblock Conll-2012 shared task: Modeling multilingual unrestricted coreference in ontonotes.
\newblock In \emph{Proceedings of the CoNLL-2012}.

\bibitem[{Radford et~al.(2019)Radford, Wu, Child et~al.}]{radford2019language}
Alec Radford, Jeff Wu, Rewon Child, et~al. 2019.
\newblock Language models are unsupervised multitask learners.
\newblock \emph{OpenAI Blog}.

\bibitem[{Raffel et~al.(2020)Raffel, Shazeer, Roberts, Lee, Narang, Matena, Zhou, Li, and Liu}]{raffel2020exploring}
Colin Raffel, Noam Shazeer, Adam Roberts, Katherine Lee, Sharan Narang, Michael Matena, Yanqi Zhou, Wei Li, and Peter~J. Liu. 2020.
\newblock Exploring the limits of transfer learning with a unified text-to-text transformer.
\newblock \emph{Journal of Machine Learning Research}, 21(140):1--67.

\bibitem[{Rahman and Ng(2012)}]{rahman2012low}
Altaf Rahman and Vincent Ng. 2012.
\newblock Coreference resolution in a low-resource language: Hindi.
\newblock In \emph{Proceedings of the 50th Annual Meeting of the Association for Computational Linguistics (Volume 2: Short Papers)}, pages 41--46.

\bibitem[{Steinberger and Jezek(2007)}]{steinberger2007text}
Josef Steinberger and Karel Jezek. 2007.
\newblock Text summarization within the information retrieval framework.
\newblock \emph{Proceedings of the 7th International Conference on Text, Speech and Dialogue}.

\bibitem[{Vilain et~al.(1995)Vilain, Burger, Aberdeen, Connolly, and Hirschman}]{vilain1995model}
Marc Vilain, John Burger, John Aberdeen, Dennis Connolly, and Lynette Hirschman. 1995.
\newblock A model-theoretic coreference scoring scheme.
\newblock In \emph{Proceedings of the 6th Message Understanding Conference (MUC-6)}, pages 45--52.

\bibitem[{Wei et~al.(2021)}]{wei2021prompt}
Jason Wei et~al. 2021.
\newblock Prompt programming for large language models: Beyond the few-shot paradigm.
\newblock In \emph{Proceedings of the 2021 Conference on Empirical Methods in Natural Language Processing}, pages 3859--3871.

\end{thebibliography}

\end{document}